\documentclass[conference]{IEEEtran}
\IEEEoverridecommandlockouts
\usepackage[bookmarks=false]{hyperref}
\usepackage{cite}
\usepackage{amsmath,amssymb,amsfonts}
\usepackage{amsthm}
\usepackage{algorithm}
\usepackage{algpseudocode}
\usepackage{graphicx}
\usepackage{textcomp}
\usepackage{subcaption}
\usepackage{xcolor}
\usepackage{hyperref}

\theoremstyle{definition}
\newtheorem{definition}{Definition}

\def\BibTeX{{\rm B\kern-.05em{\sc i\kern-.025em b}\kern-.08em
    T\kern-.1667em\lower.7ex\hbox{E}\kern-.125emX}}

\begin{document}

\title{Unfairness Discovery and Prevention For Few-Shot Regression
}


\author{\IEEEauthorblockN{Chen Zhao}
\IEEEauthorblockA{\textit{Department of Computer Science} \\
\textit{The University of Texas at Dallas}\\
Richardson Texas, USA \\
chen.zhao@utdallas.edu}
\and
\IEEEauthorblockN{Feng Chen}
\IEEEauthorblockA{\textit{Department of Computer Science} \\
\textit{The University of Texas at Dallas}\\
Richardson Texas, USA \\
feng.chen@utdallas.edu}

}

\maketitle

\begin{abstract}
We study fairness in supervised few-shot meta-learning models that are sensitive to discrimination (or bias) in historical data. A machine learning model trained based on biased data tends to make unfair predictions for users from minority groups. Although this problem has been studied before, existing methods mainly aim to detect and control the dependency effect of the protected variables (\textit{e.g.} race, gender) on target prediction based on a large amount of training data. These approaches carry two major drawbacks that (1) lacking showing a global cause-effect visualization for all variables; (2) lacking generalization of both accuracy and fairness to unseen tasks. In this work, we first discover discrimination from data using a causal Bayesian knowledge graph which not only demonstrates the dependency of the protected variable on target but also indicates causal effects between all variables. Next, we develop a novel algorithm based on risk difference in order to quantify the discriminatory influence for each protected variable in the graph. Furthermore, to protect prediction from unfairness, a fast-adapted bias-control approach in meta-learning is proposed, which efficiently mitigates statistical disparity for each task and it thus ensures independence of protected attributes on predictions based on biased and few-shot data samples. Distinct from existing meta-learning models, group unfairness of tasks are efficiently reduced by leveraging the mean difference between (un)protected groups for regression problems. Through extensive experiments on both synthetic and real-world data sets, we demonstrate that our proposed unfairness discovery and prevention approaches efficiently detect discrimination and mitigate biases on model output as well as generalize both accuracy and fairness to unseen tasks with a limited amount of training samples. 
\end{abstract}

\begin{IEEEkeywords}
causal Bayesian network, statistic parity, few-shot meta-learning, fairness generalization, bias discovery and prevention
\end{IEEEkeywords}

\section{Introduction}
Data-driven and big data technologies, nowadays, have advanced many complex domains such as healthcare, finance, social science, etc. With the development and increment of data, it is necessary to extract the potential and significant knowledge and unveil the messages hidden behind using data analysis. In data mining and machine learning, biased historical data are often learned and used to train a statistic predictive model. Depending on the application field, even though predictive models and computing process is fair, biased training data or data containing discrimination may lead to results with undesirability, inaccuracy, and even illegality. In recent years, there have been a number of news articles that discuss the concerns of bias and discrimination on crime forecasting. However, there is a lack of work that provides an extensive study on the potential bias and discrimination in public crime data and provides solutions to help produce predictive models that are free of discrimination towards the protected groups, such as African Americans. For example, 911 call data was used to predict crimes by the Seattle Police Department in 2016, but was late dropped due to potential racial bias in the provided data \cite{SeattlePolice-news-2016}.

Non-discrimination can be defined as follows: (1) people that are similar in terms of non-sensitive characteristics should receive similar predictions, and (2) differences in predictions across groups of people can only be as large as justified by non-sensitive characteristics\cite{Zliobaite-arXiv-2015}. The first condition is related to direct discrimination. For example, a hotel turns a customer away due to disability. The second condition ensures that there is no indirect discrimination, also referred to as redlining. For example, one is treated in the same way as everybody else, but it has a different and worse effect because of one's gender, race or other sensitive characters. The Equality Act \cite{TheEqualityAct-1964} calls these characters as protected characteristics. In the above-mentioned crime prediction example, even though race was not formally used as a forecasting criterion, it appeared that the geographic regions that have a much higher population of African American people have higher counts of 911 calls. Therefore, critics have voiced that human bias potentially has an influence on nowadays technology, which leads to make unfair decisions.

Machine learning models trained to give predicted outputs based on historical data will naturally inherit the past biases. With biased input, the main goal of training an unbiased model is to make the output fair. In other words, the predicted outcomes are statistically independent on protected variables (\textit{e.g.} race and gender). These may be ameliorated by attempting to make the automated decision-maker blind to some attributes. This however, is difficult, as many attributes may be correlated with the protected one \cite{Zemel-ICML-2013}. Statistical parity, also known as group fairness, ensures that the overall proportion of members in a protected group receiving predictions are identical to the proportion of the population as a whole. 

\begin{figure*}
    \centering
    \includegraphics[width = \linewidth]{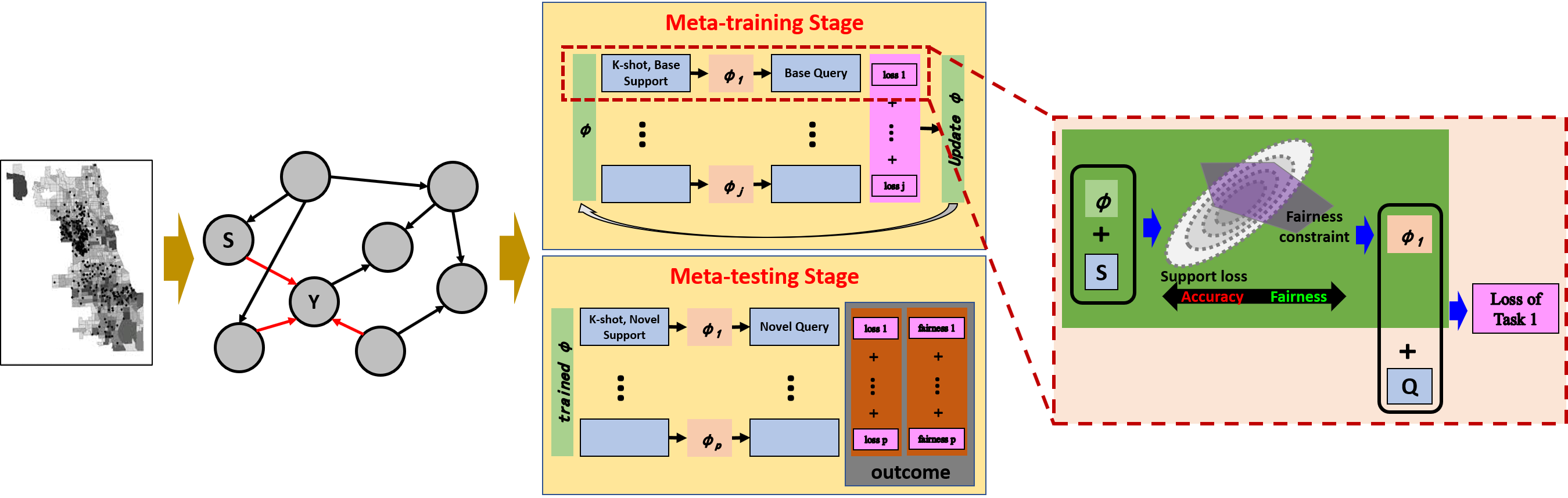}
    \caption{An overview of our proposed unfairness discovery and prevention approach in few-shot meta-learning. (Left) Discriminatory patterns in the collected data are visualized through a developed causal Bayesian network (CBN), where $S$ and $Y$ respectively denote protected attributes and target, and arrows represent causal effects between variables. (Right) A few-shot unfairness prevention approach is shown. In the meta-training stage, in each task, support loss is optimized under an unified mean-difference fairness constraint which performs a trade-off between accuracy and fairness (see the enlarged figure). The meta-parameter $\phi$ is thus iteratively optimized and then applied to calculate the final outcome, \textit{i.e.} average loss and fairness, in the meta-testing stage. $S$ and $Q$ denote the support and query data, respectively.}
    \label{fig:overview}
\vspace{-5mm}
\end{figure*}

Fairness-aware in data mining is classified into unfairness discovery and unfairness prevention. How we address unfairness detection from biased data has significant practical consequences, which will further influence the following decision-making of a machine learning model. Traditional methods for discrimination discovery from data are under the assumption that the protected variables are predefined. This does not stand up to scrutiny because one or more variables may be correlated with protected variables and have strong dependency effect on predictions. We thus, in this paper, first reveal discriminatory patterns from data through developing a causal Bayesian network (CBN) which represents a flexible useful tool in this respect as it can be used to formalize, measure, and deal with different unfairness scenarios underlying a data set. The CBN contains information of causal effect of all variables (protected and explanatory variables).

In addition, to the best of our knowledge, unfortunately, the majority of existing unfairness prevention machine learning algorithms are under the assumption of giving abundant training examples. Learning quickly, however, is another significant hallmark of human intelligence. In meta-learning, also known as learning to learn, the goal of trained model is to quickly learn a new task from a small amount of new data (\textit{i.e.} few-shot), and the model is trained by the meta-leaner to be able to learn on a large number of different tasks \cite{Finn-ICML-2017-(MAML)}. In contrast to traditional machine learning algorithms, such as multi-task learning \cite{Zhang-NSR-2018} and transfer learning \cite{Dong-ECMLPKDD-2018,Hsu-ICLR-2018}, meta-learning framework has advantages: (1) it learns across tasks where each task takes one or few samples as input; (2) it therefore efficiently speeds up model adaptation (3) and generalizes accuracy to unseen tasks. The overall idea of existing methods of meta-learning is to train a model which is capability of generalizing accuracy, rather than fairness, to unseen data tasks. But techniques for unfairness prevention and bias control in the few-shot meta-learning study are challenging and rarely touched. To ensure prediction without biases, another contribution in this paper is that we feed each support set of a task with unified group fairness constraints and minimize meta-loss overall episodes. Specifically, we mitigate biases in each episode during meta-training by controlling mean difference \cite{Calders-ICDM-2013} to a small threshold. Our experimental results based on both synthetic and real-world data sets demonstrate our approach is capability of controlling bias and decreasing loss as well as generalize both to unseen tasks.

In summary, the main contributions of this paper are listed:
\begin{enumerate}
    \item We first reveal unfairness from training data using a novel causal Bayesian network and quantify the discrimination effect of protected variables on target by developing a novel algorithm.

    \item For the first time the problem of bias control in a few-shot meta-learning regression model is introduced. Our approach efficiently mitigates the dependency of predictions on the protected attribute using mean difference from statistics.

    \item We validate the performance of our proposed approach of unfairness prevention on state-of-the-art meta-learning techniques through extensive experiments based on both synthetic and real-world data sets. Our results demonstrate the proposed approach is capability of mitigating biases and generalizing both accuracy and fairness to unseen tasks, even with minimal input.
\end{enumerate}

\section{Related Work}

In recent years, researches involving processing biased data became increasingly significant. Fairness-aware in data mining is classified into unfairness discovery and prevention. Based on the taxonomy by tasks, it can be further categorized to classification \cite{Zafar-AISTATS-2017, Hardt-NIPS-2016, Agarwal-ICML-2018, Oneto-AIES-2019, wang2019}, regression \cite{Calders-ICDM-2013,Zhao-ICDM-2019,Berk-FATML-2018,Komiyama-ICML-2018}, clustering \cite{Gondek-KDD-2005}, recommendation \cite{Singh-KDD-2018} and dimension reduction \cite{Bolukbasi-NIPS-2016}. 

Unfairness discovery aims at finding discriminatory patterns in data using data mining methods. Data mining approach for discrimination discovery typically mines association and classification rules from the data, and then assesses those rules in terms of potential discrimination \cite{Zliobaite-arXiv-2015}. A more traditional statistical approach to discrimination discovery typically fits a regression model to the data including the protected features, and then analyzes the magnitude and statistical significance of the regression coefficients at the protected attributes. If those coefficients appear to be significant, then discrimination is flagged. The existing techniques, however, only focused on the dependency of the protected attributes on target prediction, which blinded the causal effect of other explanatory variables on target and the relationship between variables. To address this flaw in unfairness detection, in this paper, we develop a causal Bayesian network containing all variables and it visually shows the causal effects among them.

Unfairness prevention develops machine learning algorithms that would produce predictive models, ensuring that those models are free from discrimination. Standard predictive models, induced by machine learning and data mining algorithms, may discriminate groups of people because (1) data bias comes from data being collected from different sources, or (2) dependence on a socially sensitive attribute was identified in the data mining community \cite{Calders-ICDM-2013}. Even though techniques for unfairness prevention on classification were well developed \cite{Zafar-AISTATS-2017, Hardt-NIPS-2016,Agarwal-ICML-2018, Oneto-AIES-2019}, limited methods have been designed for regression models and the problem on regression is more challenging. Because (1) instead of evaluating the correlation between two categorical attributes, regression aims to assess the the correlation on the categorical protected attribute and continuous target variable; (2) in classification the goal of modification led to the change of one class label into another, however in a regression task, fairness learning allows the continuous character of targets for a continuous range of potential changes. \cite{Calders-ICDM-2013} first controlled bias in a regression model by restricting the mean difference in predictions on several data strata divided using the propensity scoring method from statistics. Furthermore, \cite{Fukuchi-IEICE-2015} proposed a framework involving $\eta$-neutrality in which to use a maximum likelihood estimation for learning probabilistic models. Besides, \cite{Berk-FATML-2018,Komiyama-ICML-2018,Zhao-ICDM-2019} recently came up with convex and non-convex optimization frameworks for fairness regression. 

To the best of our knowledge, unfortunately, the majority of existing fairness-aware machine learning algorithms are under the assumption of giving abundant training examples. Learning quickly, however, is another significant hallmark of human intelligence. Several recent approaches have made significant progress in meta-learning. \cite{Vinyals-NIPS-2016-(MatchingNet)} introduced Matching Networks which employed ideas from k-nearest neighbors algorithm and metric learning based on a bidirectional Long-Short Term Memory (LSTM) to encode in the context of the support set. Prototypical networks \cite{Snell-NIPS-2017-(ProtoNet)} learn a metric space in which classification is able to be performed by computing Euclidean distances to prototype representations of each class. In addition, gradient descent based algorithms, such as  \cite{Finn-ICML-2017-(MAML),Ravi-ICLR-2017,Finn-NIPS-2018,Rusu-ICLR-2019,Antoniou-ICLR-2019}, aim to learn good model initialization so that the meta-loss is minimum. The overall idea of these state-of-the-art is to train a meta-learning model which is capability of generalizing accuracy, but less attention on fairness generalization to unseen data tasks. In this paper, our proposed approach makes up for this regret of unfairness prevention using few-shot meta-learning techniques in regression.

\section{Unfairness Discovery}
Intuitively, an attribute effects the target variable if one depends on the other. Strong dependency indicates strong effects. Currently, most fairness criteria used for evaluating and designing machine learning models focus on the relationships between the protected attribute and the system output. However, the training data can display different patterns of unfairness depending on how and why the protected attribute influences other variables. Using such criteria without fully accounting for this could be problematic. The development of technical solutions to fairness also requires considering the different, potentially intricate, ways in which unfairness can appear in the data. To this end, we construct a causal Bayesian network (CBN) using an open-source software TETRAD \cite{Tetrad-source-code}. A CBN is a graph formed by nodes representing random variables, connected by links denoting causal influence. By defining unfairness as the presence of a harmful influence from the protected attribute in the graph, CBNs provide us with a simple and intuitive visual tool for describing different possible unfairness scenarios underlying a data set. It effectively captures the existence of discrimination patterns and can provide quantitative evidence of discrimination in decision marking.

\begin{figure}[!htbp]
\captionsetup[subfigure]{aboveskip=-2pt,belowskip=-2pt}
\centering
    \begin{subfigure}[b]{0.15\textwidth}
        \includegraphics[width=\textwidth]{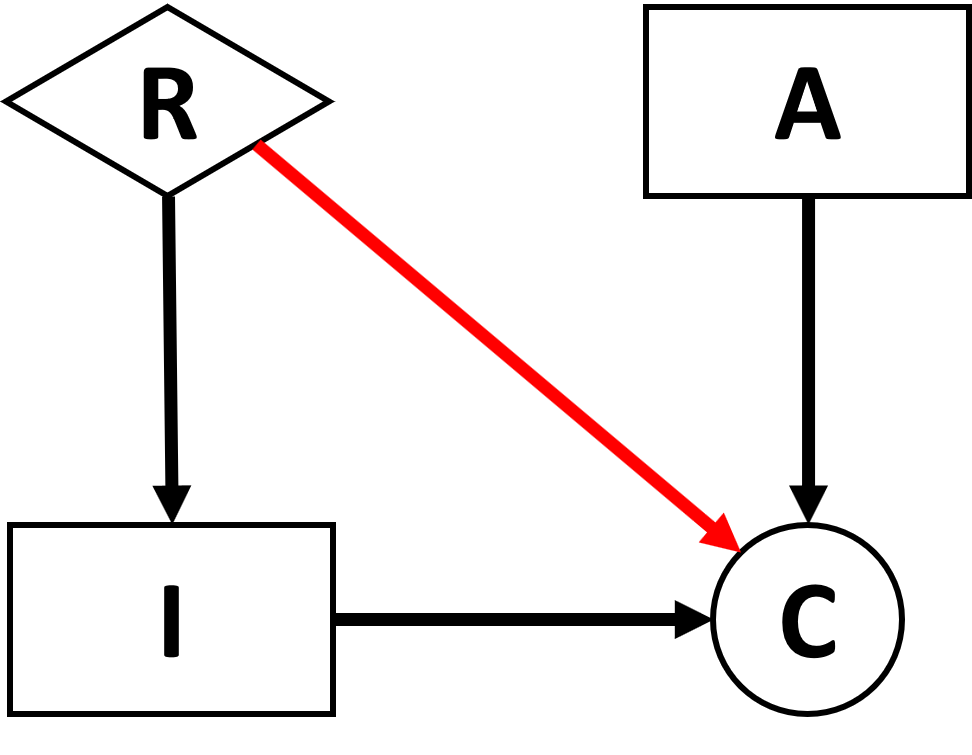}
        \caption{}
    \end{subfigure}
    \begin{subfigure}[b]{0.15\textwidth}
        \includegraphics[width=\textwidth]{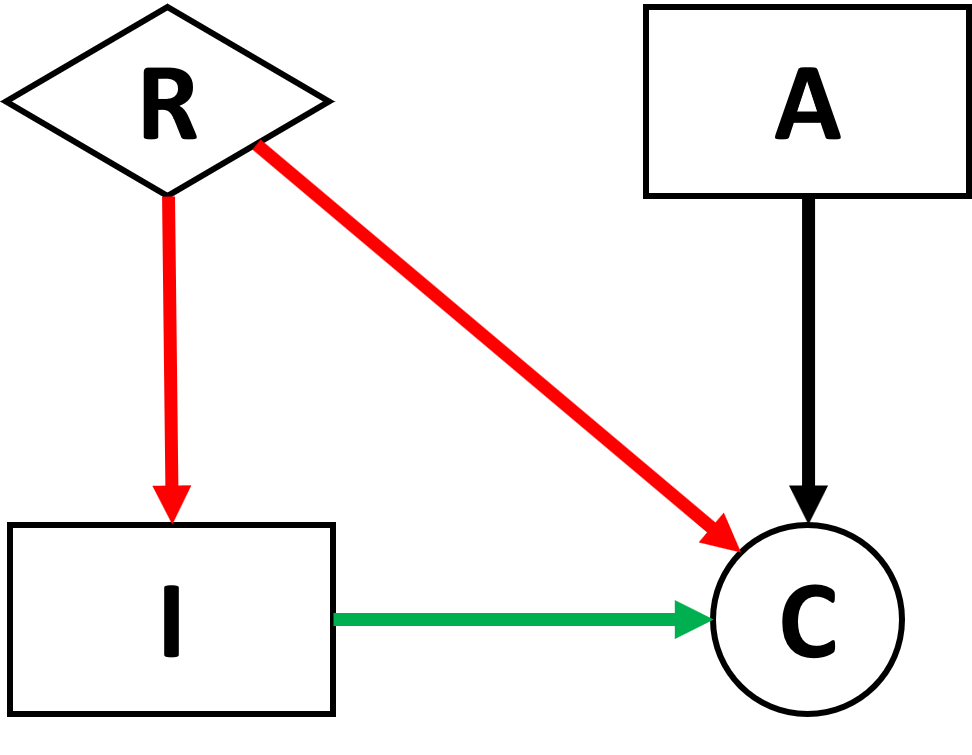}
        \caption{}
    \end{subfigure}
    \begin{subfigure}[b]{0.15\textwidth}
        \includegraphics[width=\textwidth]{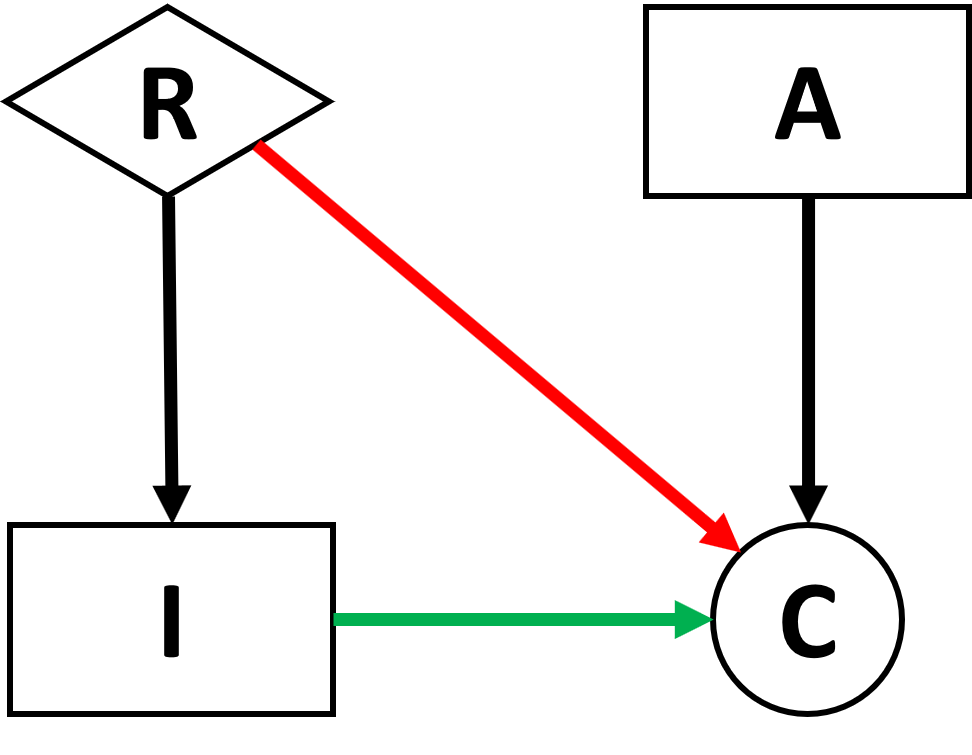}
        \caption{}
    \end{subfigure}
    \caption{CBN representing a hypothetical crime forecasting process with three possible scenarios, where red and green paths are used to indicate unfair and partially-unfair links, respectively.}
    \label{fig:CBN example}
\vspace{-2mm}
\end{figure}

Consider a hypothetical crime count prediction example in which crime are predicted based on times of being arrested ($A$), ethnicity or race ($R$), and income ($I$). The predicting process is represented by the CBN in Figure \ref{fig:CBN example}. Race has a direct influence on crime through the causal path $R\rightarrow C$ and an indirect influence through the causal path $R\rightarrow I\rightarrow C$. The direct influence captures the fact that individuals with the same arrested times who have the same income level might be treated differently based on their race (\textit{e.g.} black and non-black). The indirect influence captures differing crime counts between black and non-black individuals due to differing income levels.

\begin{definition} [Causal Path]
In a CBN, a path from node $X$ to node $Z$ is defined as a sequence of linked nodes starting at $X$ and ending at $Z$. $X$ is a cause of (has an influence on) $Z$ if there exists a causal path from $X$ to $Z$, namely a path whose links are pointing from the preceding nodes toward the following nodes in the sequence. For example, in Figure \ref{fig:CBN example}, the path $R\rightarrow I\rightarrow C$ is causal, but the path $R\rightarrow I\rightarrow C\leftarrow A$ is non causal.
\end{definition}

We depict three possible scenarios in Figure \ref{fig:CBN example}(a) to (c). In the first scenario, crime count is predicted according to the percentage of African American residents only. In the second scenario, crime counts are high in areas where African Americans percentage is high. This is because the percentage of African Americans determines the level of income and therefore $R\rightarrow I$ is consider unfair (red). As a consequence, the path $I\rightarrow C$ becomes partially unfair (green). In the third scenario, crime counts are high in the African American communities where the income level is low. In other words, for those communities with same percentage of African Americans but have high income, the crime counts may be low. This simplified example shows how CBNs can provide us with a visual framework for describing different possible unfairness scenarios. Understanding which scenario underlies a data set can be challenging or even impossible, and might require expert knowledge. It is nevertheless necessary to avoid pitfalls when evaluating or designing a decision system.

Next, to quantify the discrimination effect for each protected variable on target, we conducted the study from a data set $\mathcal{D} = \{(\mathbf{x}_i^j,y_i^j,\mathbf{s}_i^j)\}_{i=1}^{h}, j=1,...,r, i=1,...h$, where $\mathbf{x}_i^j\in\mathbb{R}^n$ denotes the $i$-th observation for the $j$-th task, $y_i^j$ denotes the corresponding numeric output, $\mathbf{s}_i^j\in\mathbb{R}^m$ represents the continuous protected attributes, and $h$ is the number of observations in each task. In order to reveal the causal effect from data, all variables, including target and the protected attributes, are discretized into three-bin categories (\textit{i.e.} low, median, high) by frequency. We considered the unfairness measure called risk difference that is denoted as 
\begin{align}
    \Delta P|_{s,j}(y|s_1, s_2) = |P(y|s_1, j) - P(y|s_2,j)|,  \quad \forall y|s,j
\end{align}
where $s_1$ and $s_2$ refer to any two different sub-populations (\textit{e.g.} $s_1$=``low" and $s_2$=``high") of a given protected variable $s\in\mathbf{s}^j$. The risk difference, $\Delta P$, thus estimates the distribution shift for target $Y$ under different protected sub-populations with a given task. Taking Crime data set as an example, each county is consider as a task and risk difference measures the distribution shift of crime rate given two sub-populations that the percentage of African American is high and low.

While the Supreme Court has resisted a ``rigid mathematical formula" defining discrimination, we adopted a generalization of the 80 percent rule advocated by the US Equal Employment Opportunity Commission (EEOC)\footnote{https://www.eeoc.gov/}. Formally, we define a fairness constraint of discrimination based on statistical parity:
\begin{align}
    P(P|_{j}(y|s)\leq \epsilon) \leq 80\%
\end{align}
where $\epsilon$ is a given threshold to account for a degree of randomness in the decision making process and sampling. We adopted the value $\epsilon=0.05$ as used in \cite{Zhang-SBP-BRiMS-2016}. Key steps of calculating risk difference are described in Algorithm \ref{alg:RiskDiference}.

\begin{algorithm}[h]
\caption{Unfairness Discovery Using Risk Difference.}
\textbf{Require: } All variables are discretized into three-category bin.
\begin{algorithmic}[1]
\State Initialization $n=0, m=0$
\For{each task $j$}
    \For{each configuration $y$ of $Y$}
        \For{each different configurations $\{s_1, s_2\}$ of $S$}
            \State Calculate risk difference: $\Delta P|_{s,j}(y|s_1, s_2) = |P(y|s_1, j) - P(y|s_2,j)|$
            \If{$\Delta P|_{s,j}(y|s_1, s_2)\leq \epsilon$}
                \State $n=n+1$
            \EndIf
            \State $m=m+1$
        \EndFor
    \EndFor
\EndFor
\State Calculate ratio: $r = n/m$.
\State \Return The ratio $r$.
\end{algorithmic}
\label{alg:RiskDiference}
\end{algorithm}
\vspace{-5mm}

\section{Unfairness Prevention in Few-Shot Regression}
Unfairness prevention develops machine learning algorithms that would produce predictive models, ensuring that those models are free from discrimination. In contrast to traditional regression settings, in this section, we introduce a novel few-shot discrimination prevention learning model based on the Model Agnostic Meta-Learning (MAML) framework \cite{Finn-ICML-2017-(MAML)}, which is able to quickly learn a new task from a small amount of data (\textit{i.e.} $K$-shot) and to generalize fairness onto unseen tasks. For simplicity, we select the most representative protected variable $s\in\mathbf{s}$ which takes the greatest risk difference and convert it into a binary variable $\{s_+,s_-\}$. However, our ideas can be easily extended to many protected attributes with multiple levels.

\subsection{Evaluation of Statistical Parity}
A learning model is considered illegal discrimination if it has a disproportionately adverse effect on members of a protected group (\textit{e.g.} race, gender). In other words, statistical parity ensures that the overall proportion of members in a protected group receiving prediction is identical to the proportion of the population as a whole.

To formulate, we split the data set for each task $j$ into an episode $\{\mathcal{D}_j^\mathcal{S}, \mathcal{D}_j^\mathcal{Q}\}$, where $\mathcal{D}_j^\mathcal{S}$ and $\mathcal{D}_j^\mathcal{Q}$ are support and query set, respectively. A $K$-shot learning set $\mathcal{D}_j=\{\mathbf{x}_i, y_i, s_i\}_{i=1}^K$, where $s_i\in\{s_+,s_-\}$ is the binary protected attribute. For regression problem, since $y_i\in\mathbb{R}$ is a numeric scalar, it is not possible to use typical same-type measures of dependency like correlation coefficient and point-wise mutual information to quantify the statistical dependency between $s$ and $y$. To quantify the effect of protected attribute $s$ on its target $y$ for $\mathcal{D}_j$, we apply Mean Difference (MD) for evaluating biases in regression problems. 

\theoremstyle{definition}
\begin{definition}[Mean Difference]
The mean difference (MD) of numeric target variable $y$ in data set $\mathcal{D}$, partition into $\mathcal{D}_+$ and $\mathcal{D}_-$ by a binary protected variable $s$ is given by:
\begin{align*}
    MD (y, s; \mathcal{D}) = |\frac{\sum_{(\mathbf{x},y)\in\mathcal{D}_+} y}{N_+} - \frac{\sum_{(\mathbf{x},y)\in\mathcal{D}_-} y}{N_-}|
\end{align*}
where $N_+$ and $N_-$ are group sizes of $\mathcal{D}_+$ and $\mathcal{D}_-$. The mean difference is a positive number with a value of zero indicating no dependency of target on the protected variable.
\end{definition}

\subsection{Fair Model Agnostic Few-Shot Meta-Learning}
Meta-learning is also known as learning to learn. In a general meta-learning setting, it consists of meta-train and meta-test stages where each contains a number of mini-batches of episodes split into support and query sets. We consider a distribution over tasks $p(\mathcal{T})$ that we want our model to be able to adapt to. In a $K$-shot learning setting, a task $\mathcal{T}_j$ is sampled from $p(\mathcal{T})$, where the subscript $j$ represents the $j$-th task of a mini-batch. In the supervised learning setting, supposing the meta-model is a parameterized function $f_\phi$ with parameters $\phi$. In a general meta-learning model, the goal is to learn an optimized meta-parameter $\phi$ so that the summation of query losses $l_{\mathcal{T}_j}(f_\phi)$ over all meta-training tasks is minimum.

\begin{align}
    \phi^* = \arg\min_\phi \mathbb{E}_{\mathcal{T}\sim p(\mathcal{T})} l_{\mathcal{T}}(f_\phi)
\end{align}

The use of only $K$ training examples for learning a new task is often referred to as $K$-shot learning. During meta-training, $\phi$ is updated iteratively. The trained meta-model is evaluated through a set of tasks that are not included in the meta-training procedure. To formulate the supervised regression problem in the context of the meta-learning model, the loss functions, mean squared error, is applied. It is represented by the error between the model's output for $\mathbf{x}$ and the corresponding target $\mathbf{y}$. In order to control biases in prediction, it is required to restrict a statistical parity score $g_{\mathcal{T}_j}(f_\phi)$ for each task with a user-defined fairness threshold $c>0$. The objective function of a single task takes the form:

\begin{align}
\label{regression_prob}
    &\textbf{minimize}_{\phi_j} \quad l_{\mathcal{T}_j}(f_\phi) = \sum_{\mathbf{x}_i, \mathbf{y}_i\sim\mathcal{T}_j}||f(\mathbf{x}_i;\phi)-\mathbf{y}_i||_2^2\\
    &\textbf{subject to} \quad g_{\mathcal{T}_j}(f_\phi)\leq c\nonumber
\end{align}

where $\mathbf{x}_i, \mathbf{y}_i$ are a pair of input feature vector and output target sampled from task $\mathcal{T}_j$, $c$ is a small positive fairness relaxation, and $g_{\mathcal{T}_j}(f_\phi)$ is the mean difference of the continuous prediction bounded by $c$.

\begin{align}
    g_{\mathcal{T}_j}(f_\phi)= \Bigg| \frac{\sum_{(\mathbf{x}_i, \mathbf{y}_i\sim \mathcal{T}_j)\in\mathcal{D}_+} f(\mathbf{x}_i;\phi)}{N_+} - \\
    \frac{\sum_{(\mathbf{x}_i, \mathbf{y}_i\sim \mathcal{T}_j)\in\mathcal{D}_-} f(\mathbf{x}_i;\phi)}{N_-} \Bigg| \nonumber
\end{align}

To solve the optimization problem, we thus introduce an unified Lagrange multiplier $\lambda\geq 0$ for all tasks and the Lagrange function $\mathcal{L}_{\mathcal{T}_j}(\phi, \lambda)$  of a single task is defined by 
\begin{align}
    \mathcal{L}_{\mathcal{T}_j}(\phi, \lambda) = l_{\mathcal{T}_j}(f_\phi) + \lambda(g_{\mathcal{T}_j}(f_\phi)-c)
\end{align}
Therefore the original problem can be finally seen by minimizing $\mathcal{L}_{\mathcal{T}_j}(\phi, \lambda)$ for each task and thus mitigates dependency of prediction on the protected attribute. 

The goal of training a single task is to output a local parameter $\phi_j$ given the meta-parameter $\phi$ such that it minimizes the task loss $l_{\mathcal{T}_j}(f_\phi)$ subject to the task constraint $g_{\mathcal{T}_j}(f_\phi)\leq c$. Next, to update the meta-parameter, we minimize the generalization error $\mathcal{L}^{meta}$ using query sets across every task in the batch such that the query constraints are satisfied. Formally, the learning objective across all tasks is

\begin{align}
\label{meta_problem}
    \min_{\phi} \quad &\mathcal{L}^{meta}(\sum_{j=1}^T\mathcal{D}_j^Q,\phi) =\sum_{j=1}^T l_{\mathcal{T}_j}(f_{\phi_j})(\mathcal{D}^Q_j,\phi_j)
\end{align}

where $\phi_j=\arg\min_{\phi_j, g_{\mathcal{T}_j}(f_\phi)\leq c}l_{\mathcal{T}_j}(f_\phi)$ is the local optimum for each task. A step-by-step learning algorithm for the unfairness prevention approach in few-shot regression is proposed in Algorithm \ref{alg:bias prevention few-shot}.

\begin{algorithm}[h]
\caption{Unfairness Prevention in Few-Shot Regression.}
\textbf{Require: } $p(\mathcal{T})$: distribution over tasks.\\
\textbf{Require: } $\alpha, \beta$: step size hyperparameters.\\
\textbf{Require: } $q$: inner gradient update steps.
\begin{algorithmic}[1]
\State Randomly initialize $\phi$
\While{not done}
    \State Sample batch of tasks $\mathcal{T}_j$
    \For{all $\mathcal{T}_j = \{\mathcal{D}_j^\mathcal{S}, \mathcal{D}_j^\mathcal{Q}\}$}
        \State Sample $K$ datapoints from $\mathcal{D}_j^\mathcal{S}=\{\mathbf{x}^j, y^j, s^j\}$
        \State $\phi_j \leftarrow \phi$
        \For{$q=1,2,...$}
            \State Evaluate $\nabla_{\phi_j} \mathcal{L}_{\mathcal{T}_j}(\phi_j, \lambda)$ using $\mathcal{D}_j^\mathcal{S}$
            \State $\phi_j \leftarrow \phi_j-\alpha \nabla_{\phi_j} \mathcal{L}_{\mathcal{T}_j}(\phi_j, \lambda)$ 
        \EndFor
        \State Sample datapoints from $\mathcal{D}_j^\mathcal{Q}=\{\mathbf{x}^j, y^j, s^j\}$
        \State Evaluate query loss $l_{\mathcal{T}_j}(f_{\phi_j})$ and query fairness $g_{\mathcal{T}_j}(f_{\phi_j})$ using $\mathcal{D}_j^\mathcal{Q}$
    \EndFor
    \State Update $\phi \leftarrow \phi - \beta\nabla_\phi\sum_{\mathcal{T}_j\sim p(\mathcal{T})} l_{\mathcal{T}_j}(f_{\phi_j})$
    \State Evaluate training fairness $mean(g_{\mathcal{T}_j}(f_{\phi_j}))$
\EndWhile
\end{algorithmic}
\label{alg:bias prevention few-shot}
\end{algorithm}

\section{Experiments}
In the section, we first demonstrate the individual utility of unfairness discovery approach that introduced in section III based on Crime data set. Then we conduct extensive experiments to validate the proposed few-shot unfairness prevention algorithm on both synthetic and real-world data sets. 

\subsection{Unfairness Discovery from Crime Data Set}
\textbf{Chicago Communities and Crime} data set \cite{Zhao-ICDM-2019} includes information relevant to crime (\textit{e.g.} household, unemployment) as well as demographic information (such as race and gender) in different communities across the Chicago city in 2015. These information were separately collected from \textit{American FactFinder (AFF)} \footnote{https://factfinder.census.gov/faces/nav/jsf/pages/index.xhtml} which is an online and self-service database provided by the U.S. Census Bureau and then aggregated various sources to the final data prepared for experiments. More specifically, the economy related information such as points of interests (such as businesses and attractions) was extracted from location-based social networks (Foursquare check-in data). While the crime and geographical information in the data correspond to the specific crimes that occurred, our investigations clearly indicate that the local neighborhood information can provide strong indication about future crimes. In this data set, resident population of various ethnicities are considered as protected variables and crime count is the target attribute that we need to detect discrimination from. 

\begin{figure}[!htbp]
    \centering
    \includegraphics[width=\linewidth]{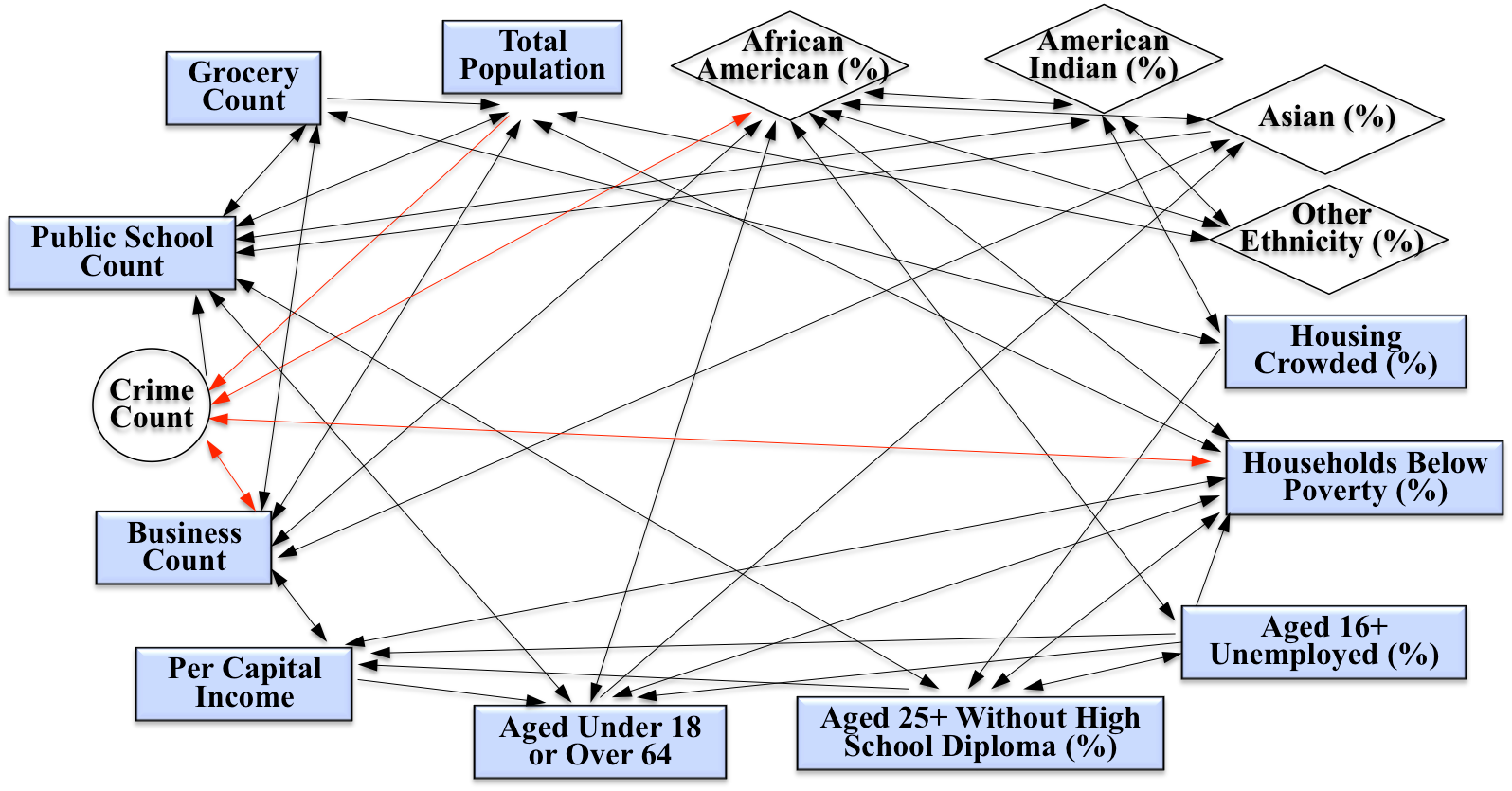}
    \caption{Causal Bayesian network conducted using the Crime data set. Red arrows are highlighted the causal effect between variables that target (crime count) is dependent on. }
    \label{fig:CBN-crime}
\end{figure}

\textbf{Experimental Results.} First, we reveal unfairness from the Crime data set by conducting the causal Bayesian network (CBN) shown in Figure \ref{fig:CBN-crime}, where \textit{``African American (\%)"} refers to the percentage of African American residents in each of census tract (geo-location unit) and \textit{``Other Ethnicity (\%)"} is the percentage of ethnicity groups other than white, African American, Asian, and American Indian groups. In the result of CBN, the biases and discrimination are modeled based on the causal paths from one variable to another. As shown in Figure \ref{fig:CBN-crime}, highlighted paths demonstrate that crime count is dependent on four variables, \textit{``Total Population", ``African American (\%)", ``Households Below Poverty (\%)",} and \textit{``Business Count"}. We thus need to consider a correct partition of the CBN network, in order to suppress all other influences rather than discrimination. Some of these are spurious and some although causal, can be explained by other attributes and hence are not regarded as discrimination. 

\begin{table}[!h]
\normalsize
    \centering
    \caption{Analysis of statistical parity on protected groups for the Crime data set.
    }
    \begin{tabular}{c|c}
        \hline
        \textbf{Protected Groups (\%)} & \textbf{Unfairness} \\
        \hline
        African American & $22.20\%$ \\
        \hline
        American Indian & $46.91\%$ \\
        \hline
        Asian & $40.70\%$\\
        \hline
        Other Ethnicity & $40.70\%$ \\
        \hline
    \end{tabular}
    \label{tab:statistical parity measures}
\vspace{-3mm}
\end{table}

To quantify dependency effect of protected variable on crime count, we block all causal paths from the protected attribute $s$ to target $y$ in the CBN and analyze each protected variables using statistical parity approach introduced in section III. The estimated statistical parity measures of different protected groups are shown in Table \ref{tab:statistical parity measures}. The result indicates that the Crime data contains potential bias based on the $80\%$ rule. A lower statistical parity indicates a higher unfairness causal effect of crime count towards the protected group.

\begin{figure}[!htbp]
\captionsetup[subfigure]{aboveskip=-2pt,belowskip=-2pt}
\centering
    \begin{subfigure}[b]{0.235\textwidth}
        \includegraphics[width=\textwidth]{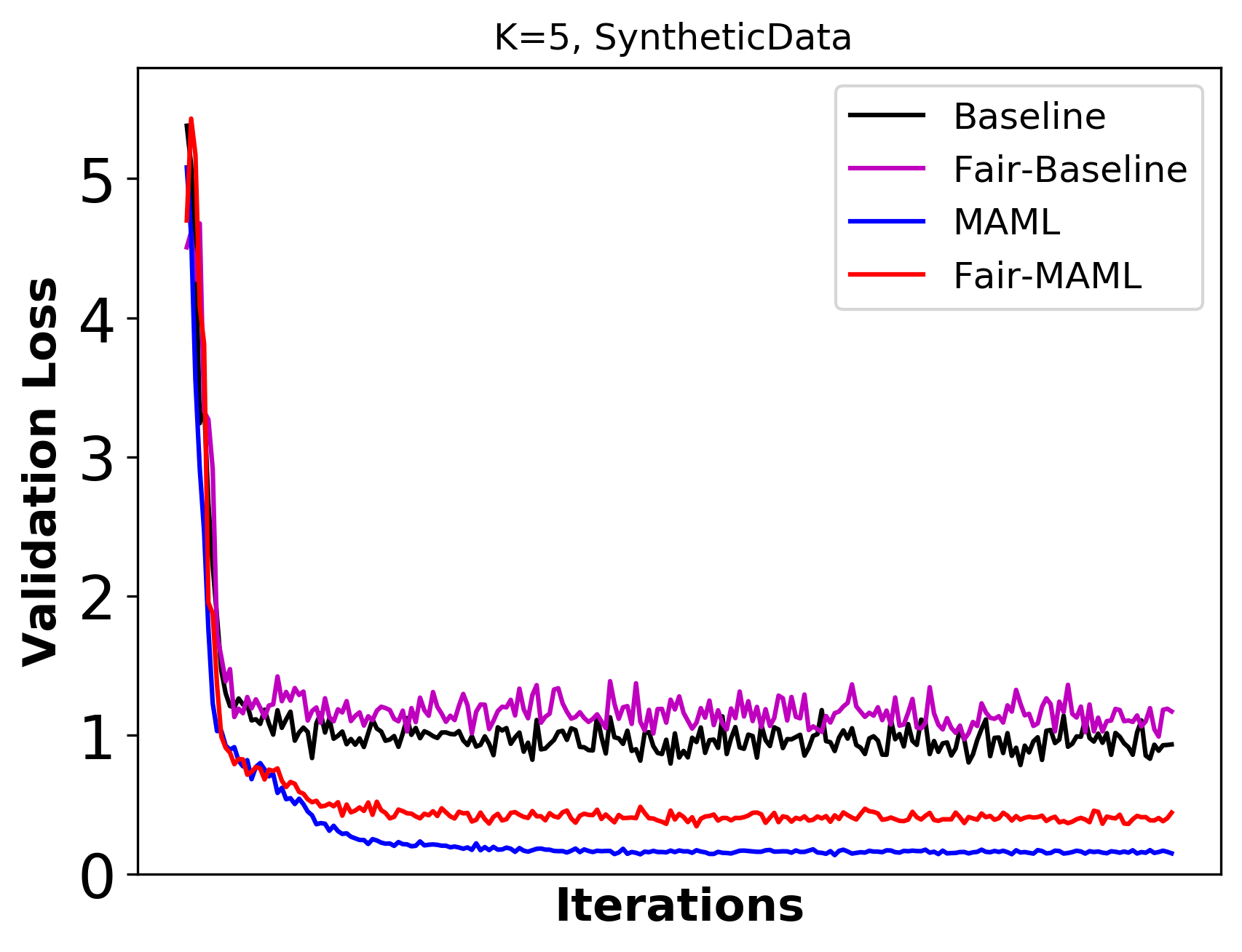}
        \caption{}
    \end{subfigure}
    \begin{subfigure}[b]{0.235\textwidth}
        \includegraphics[width=\textwidth]{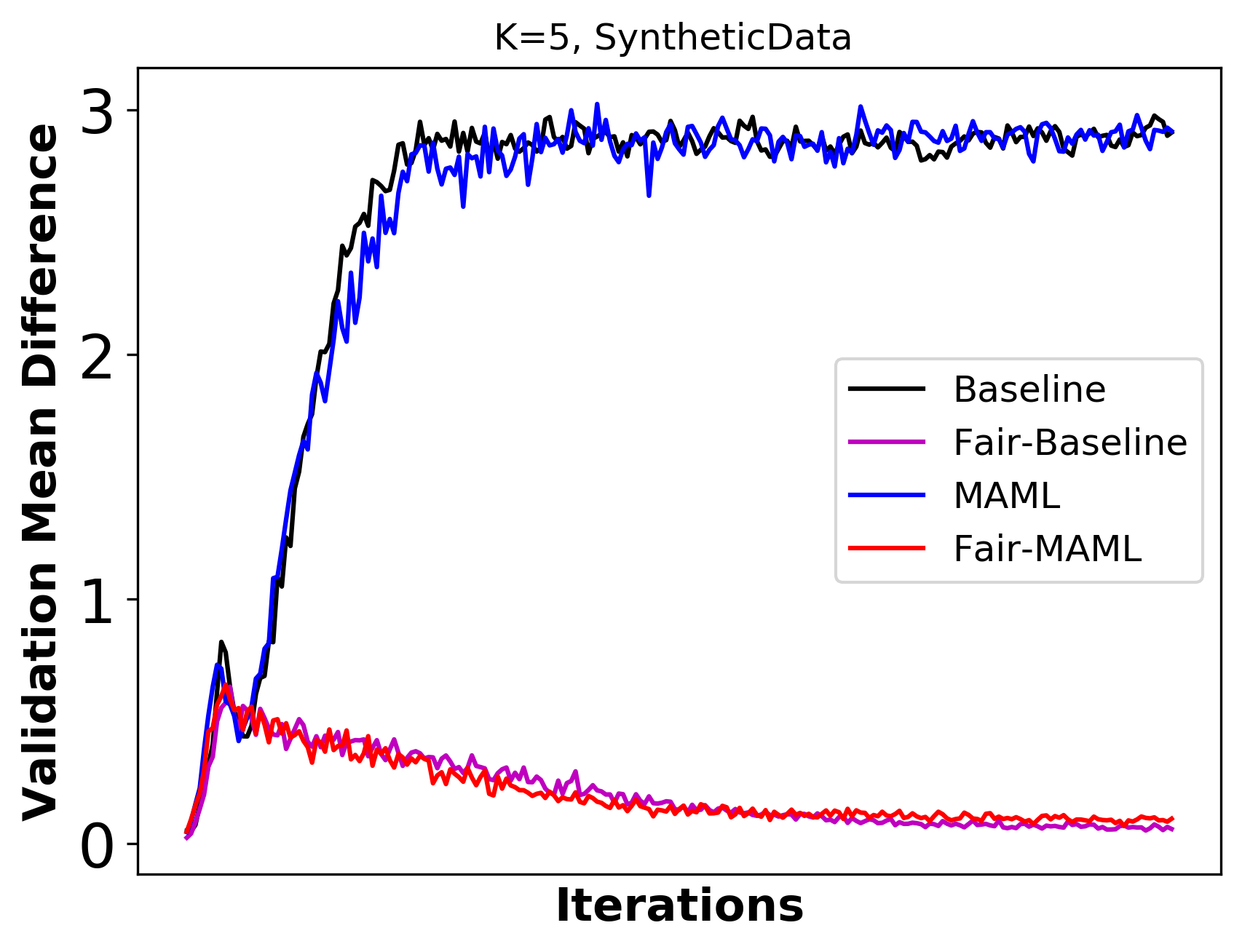}
        \caption{}
    \end{subfigure}
    
    \begin{subfigure}[b]{0.235\textwidth}
        \includegraphics[width=\textwidth]{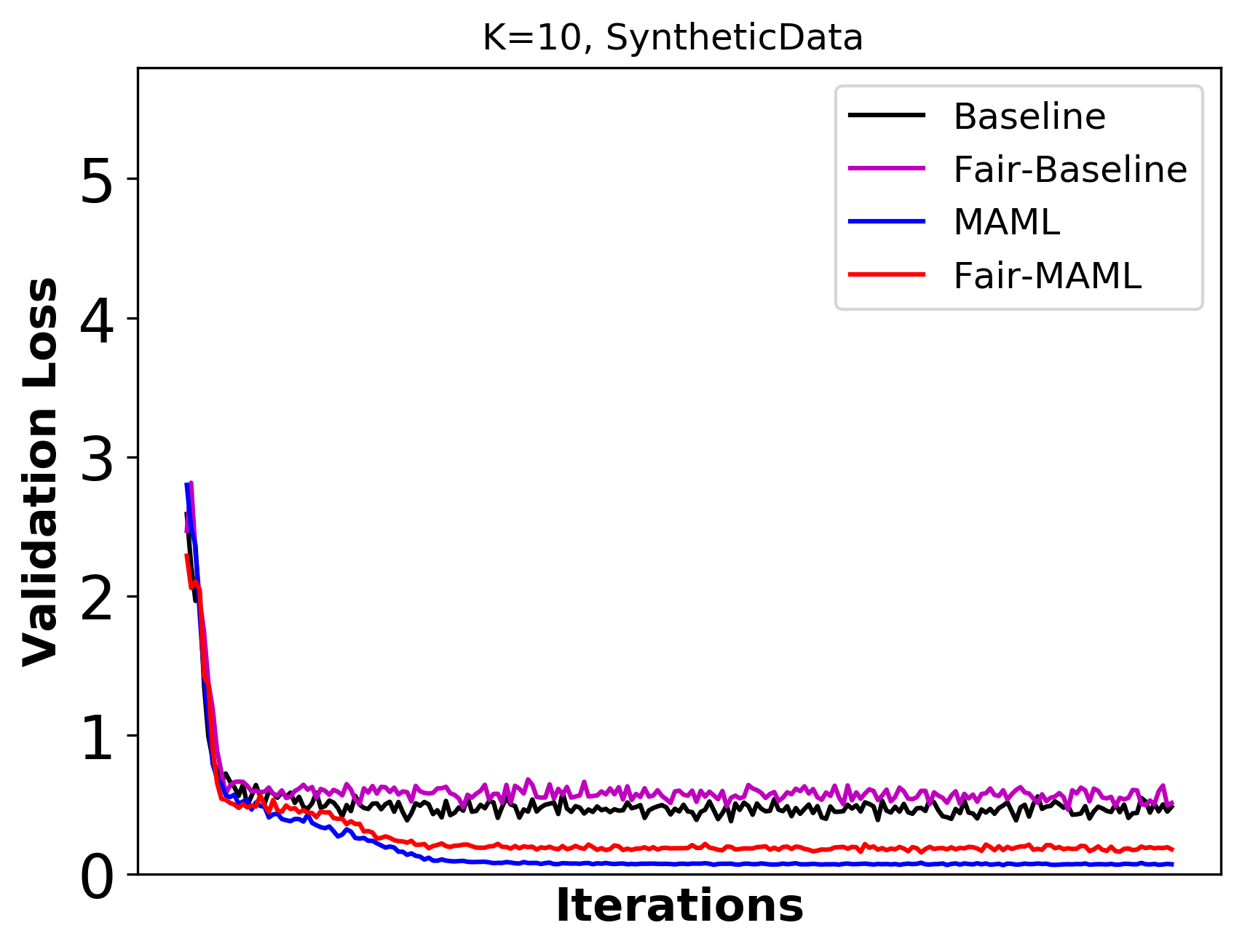}
        \caption{}
    \end{subfigure}
    \begin{subfigure}[b]{0.235\textwidth}
        \includegraphics[width=\textwidth]{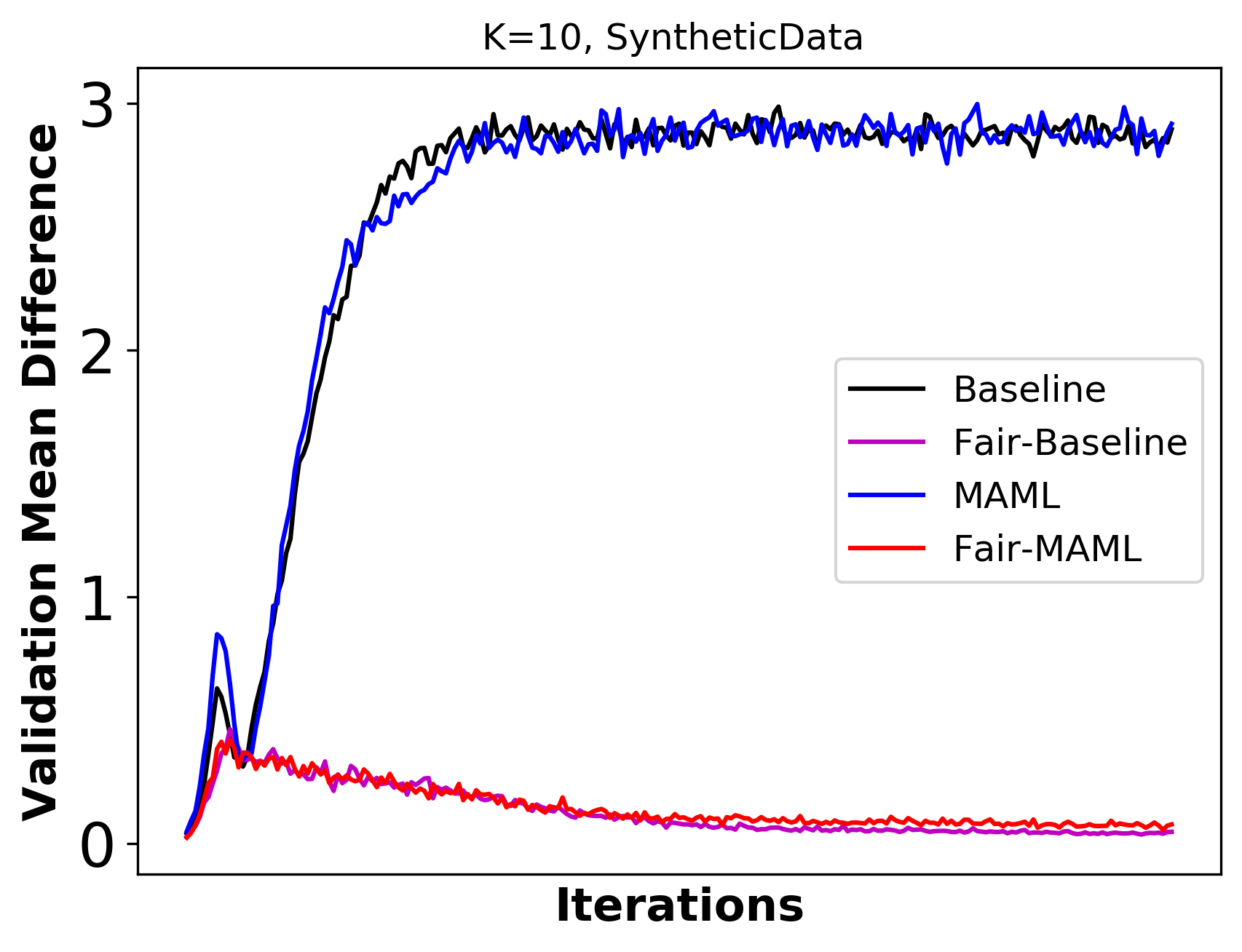}
        \caption{}
    \end{subfigure}
    
    \begin{subfigure}[b]{0.235\textwidth}
        \includegraphics[width=\textwidth]{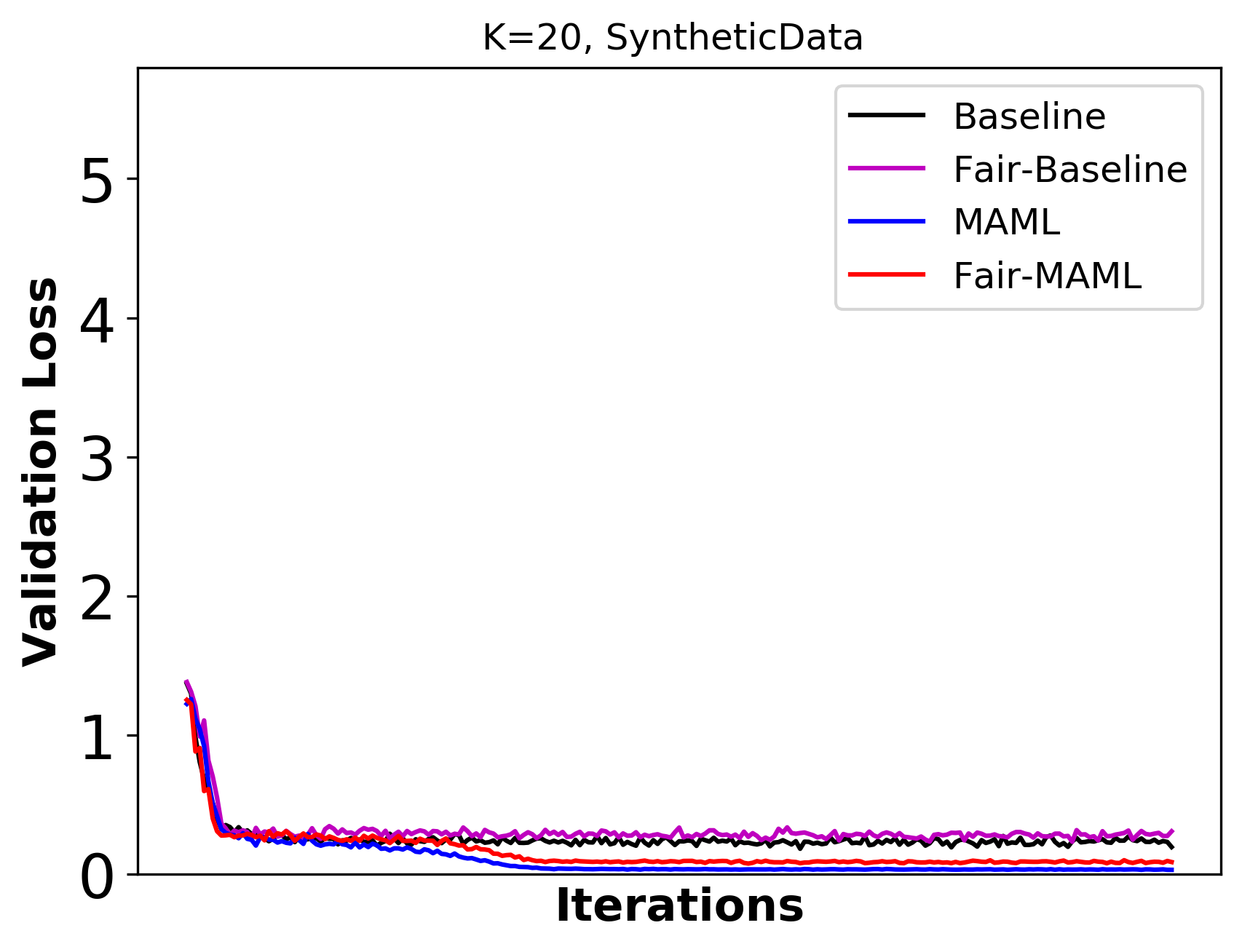}
        \caption{}
    \end{subfigure}
    \begin{subfigure}[b]{0.235\textwidth}
        \includegraphics[width=\textwidth]{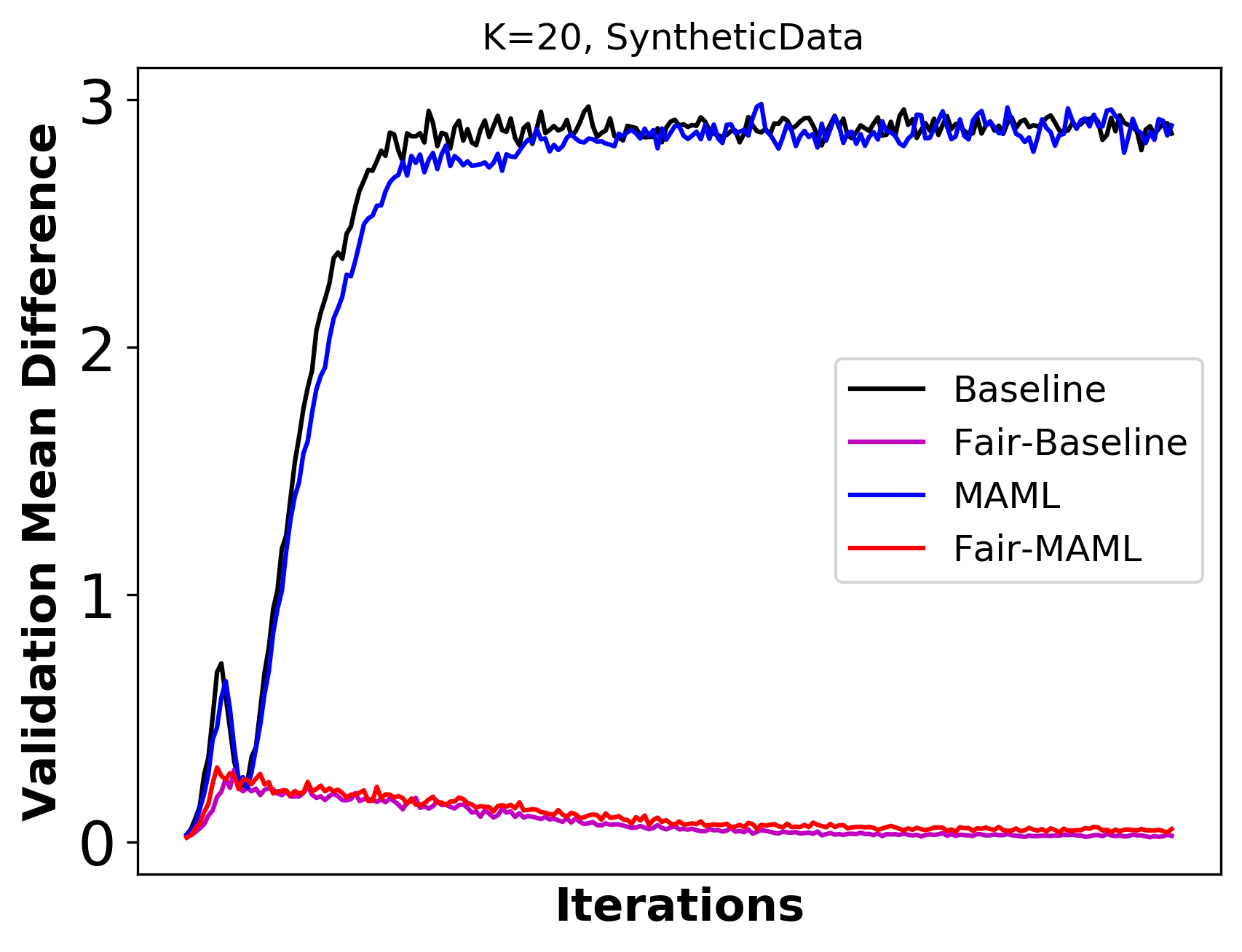}
        \caption{}
    \end{subfigure}
    
    \caption{Validation loss and mean difference stabilization over iterations.}
    \label{fig:syn loss/md iterations}
\vspace{-5mm}
\end{figure}


\subsection{Unfairness Prevention with Few-Shot Learning}

\begin{figure*}[!htbp]
\captionsetup[subfigure]{aboveskip=-2pt,belowskip=-2pt}
\centering
    \begin{subfigure}[b]{0.245\textwidth}
        \includegraphics[width=\textwidth]{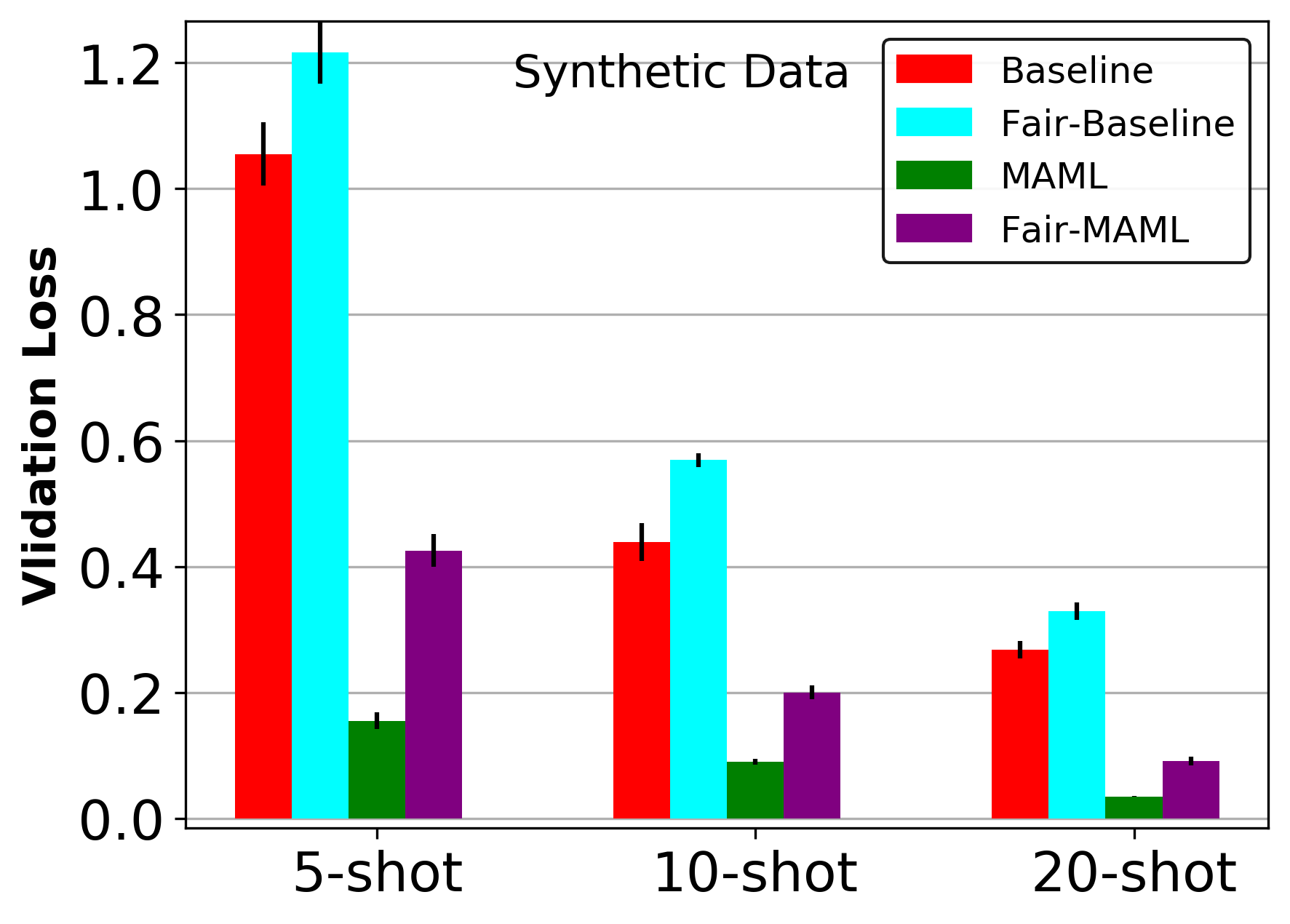}
        \caption{}
    \end{subfigure}
    \begin{subfigure}[b]{0.245\textwidth}
        \includegraphics[width=\textwidth]{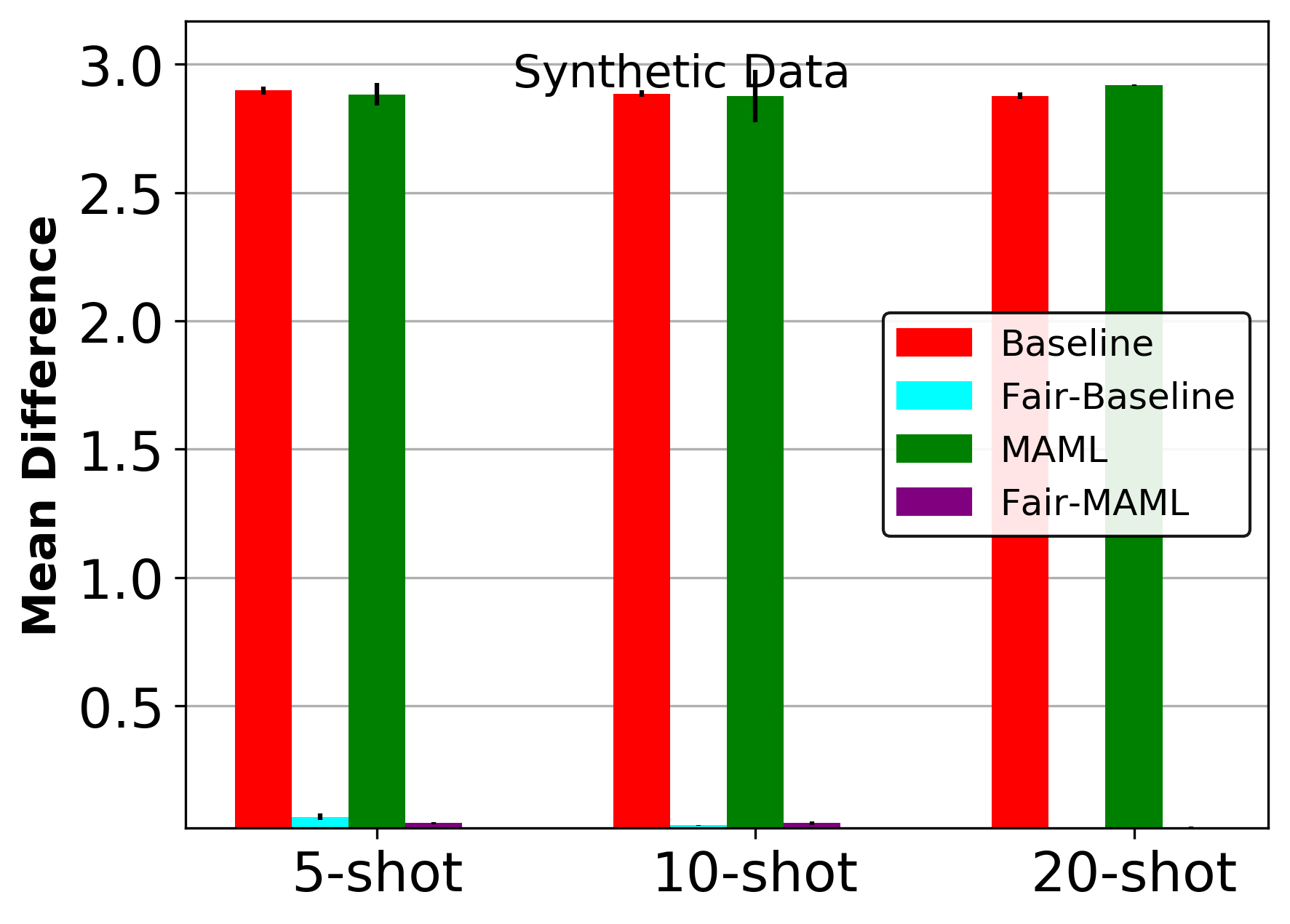}
        \caption{}
    \end{subfigure}
    \begin{subfigure}[b]{0.245\textwidth}
        \includegraphics[width=\textwidth]{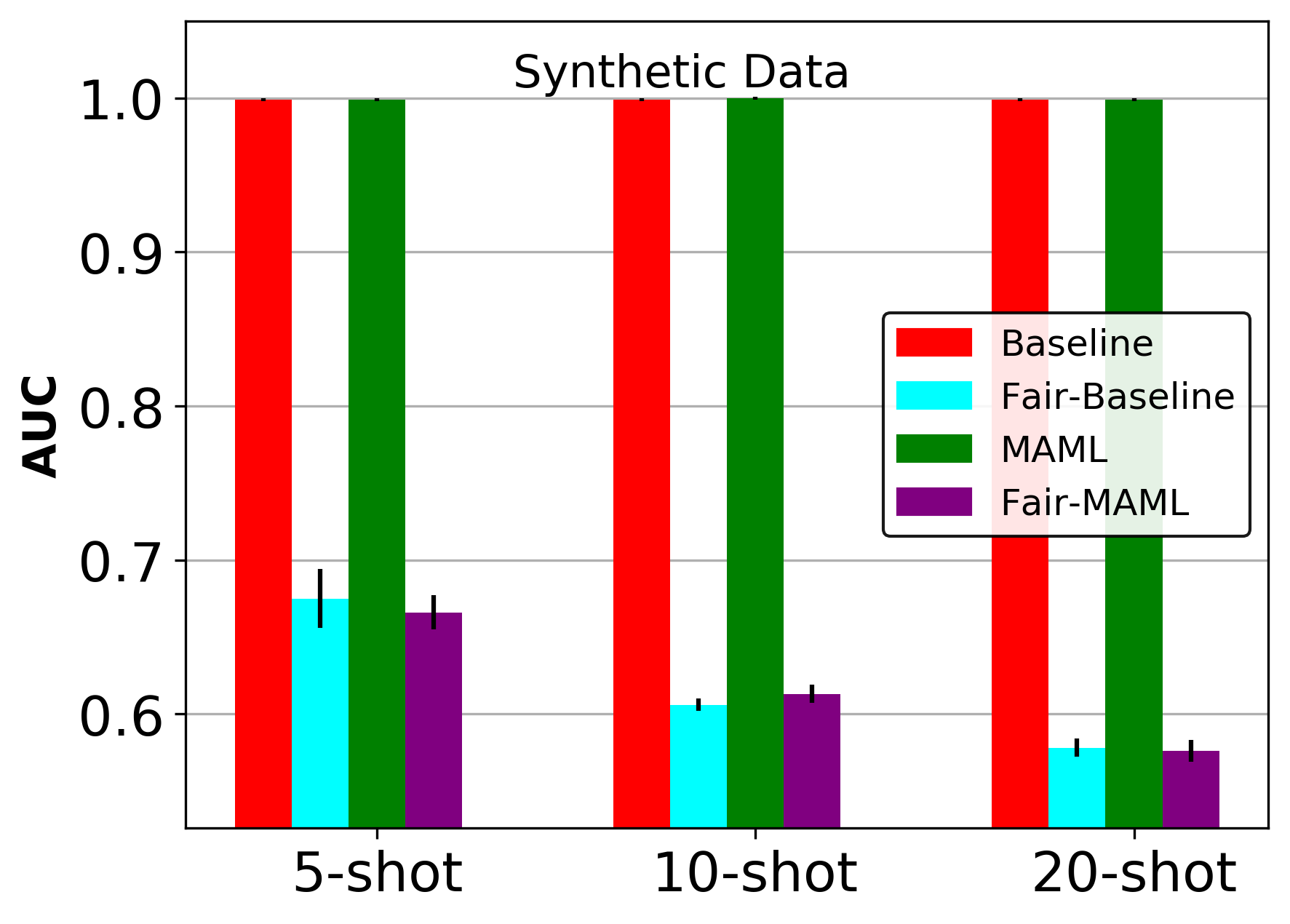}
        \caption{}
    \end{subfigure}
    \begin{subfigure}[b]{0.245\textwidth}
        \includegraphics[width=\textwidth]{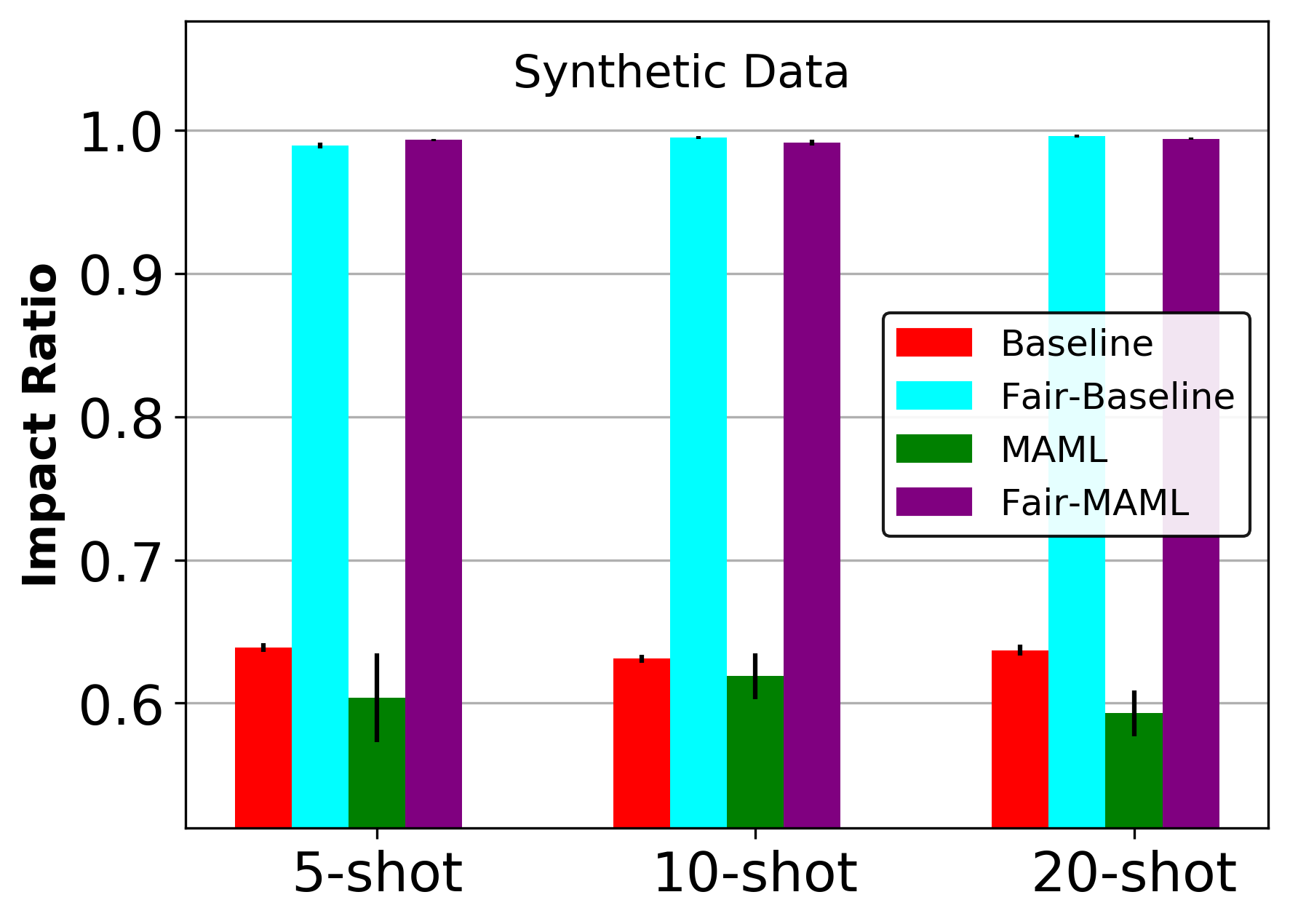}
        \caption{}
    \end{subfigure}

    \begin{subfigure}[b]{0.245\textwidth}
        \includegraphics[width=\textwidth]{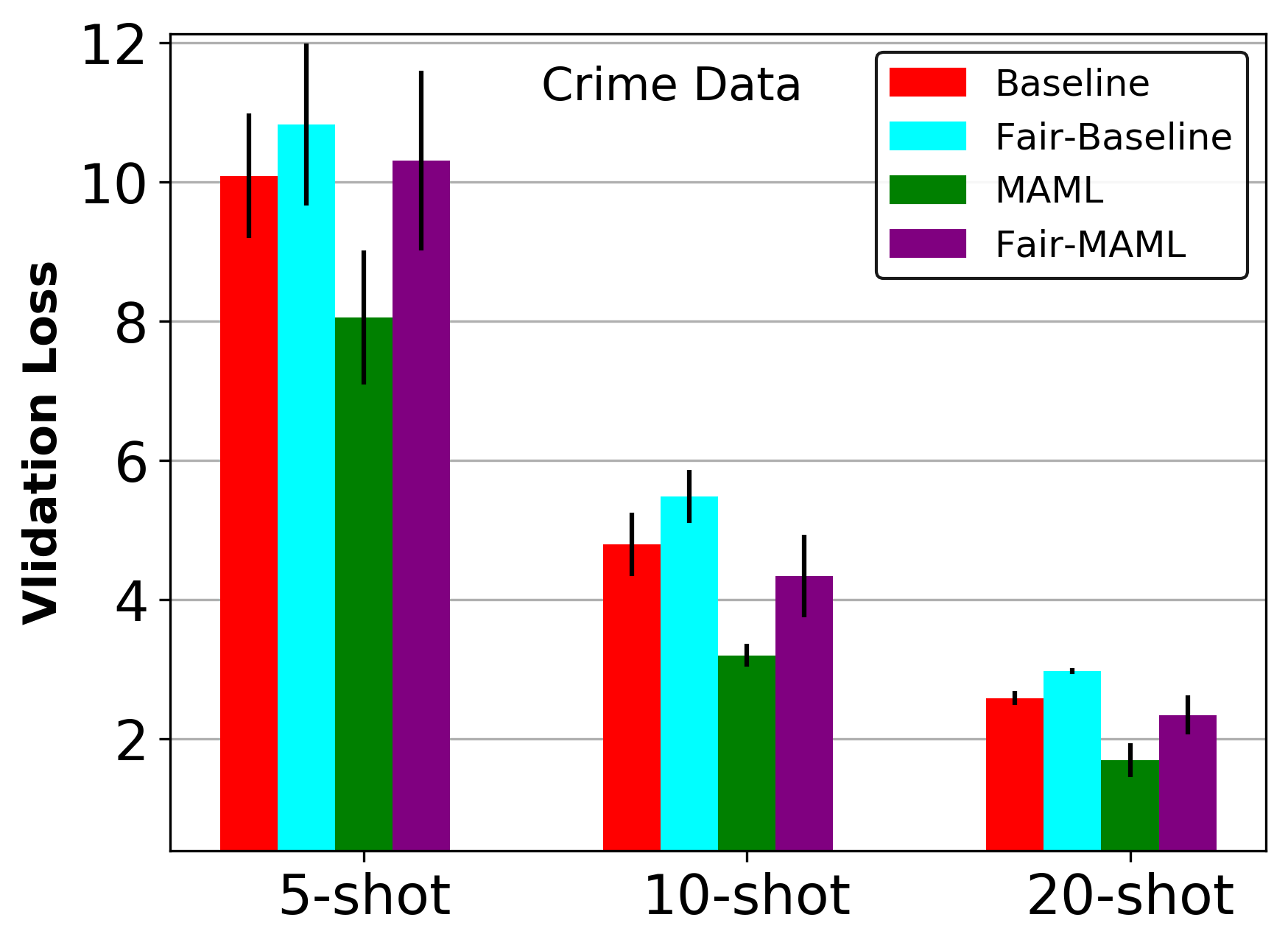}
        \caption{}
    \end{subfigure}
    \begin{subfigure}[b]{0.245\textwidth}
        \includegraphics[width=\textwidth]{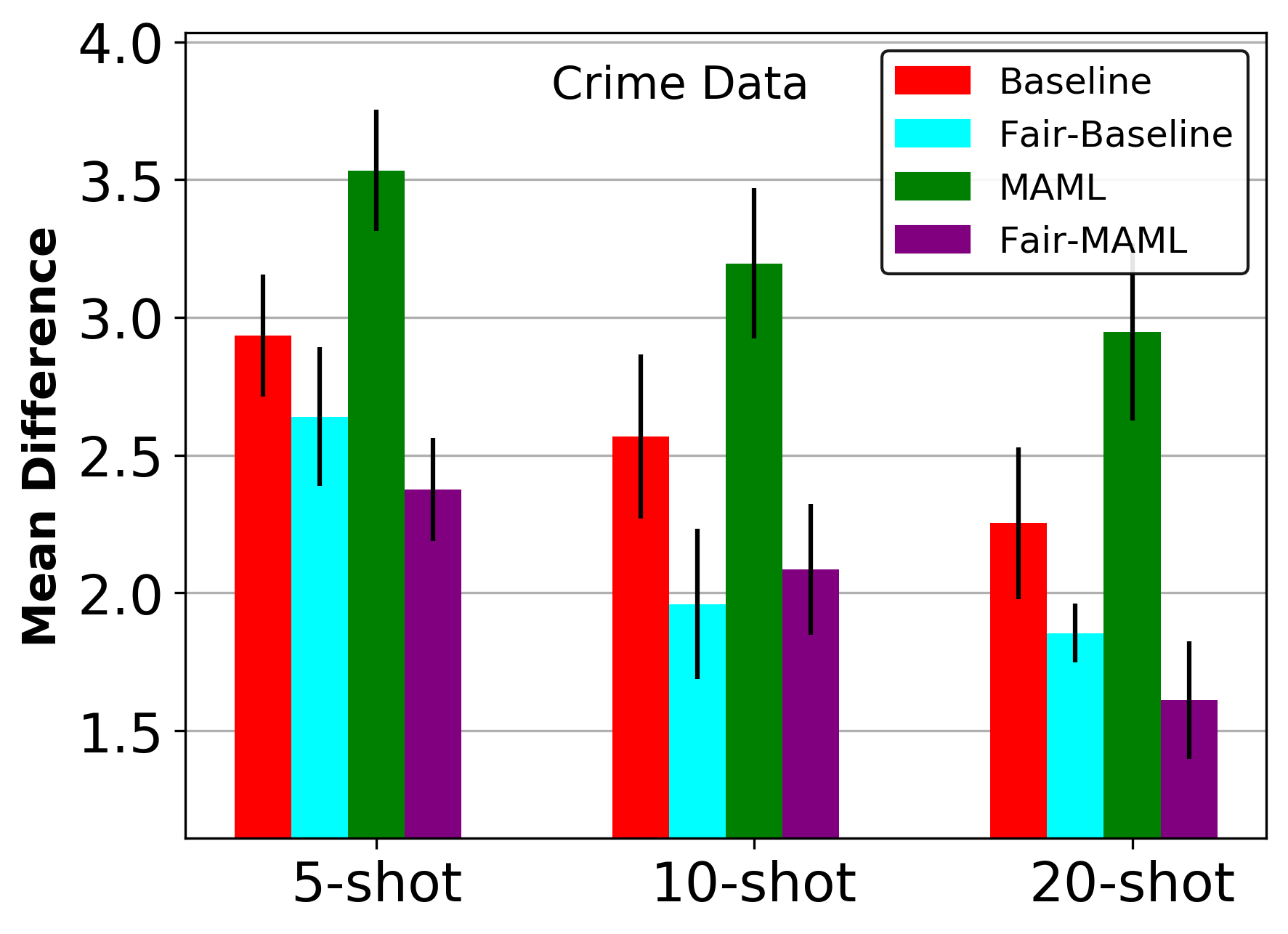}
        \caption{}
    \end{subfigure}
    \begin{subfigure}[b]{0.245\textwidth}
        \includegraphics[width=\textwidth]{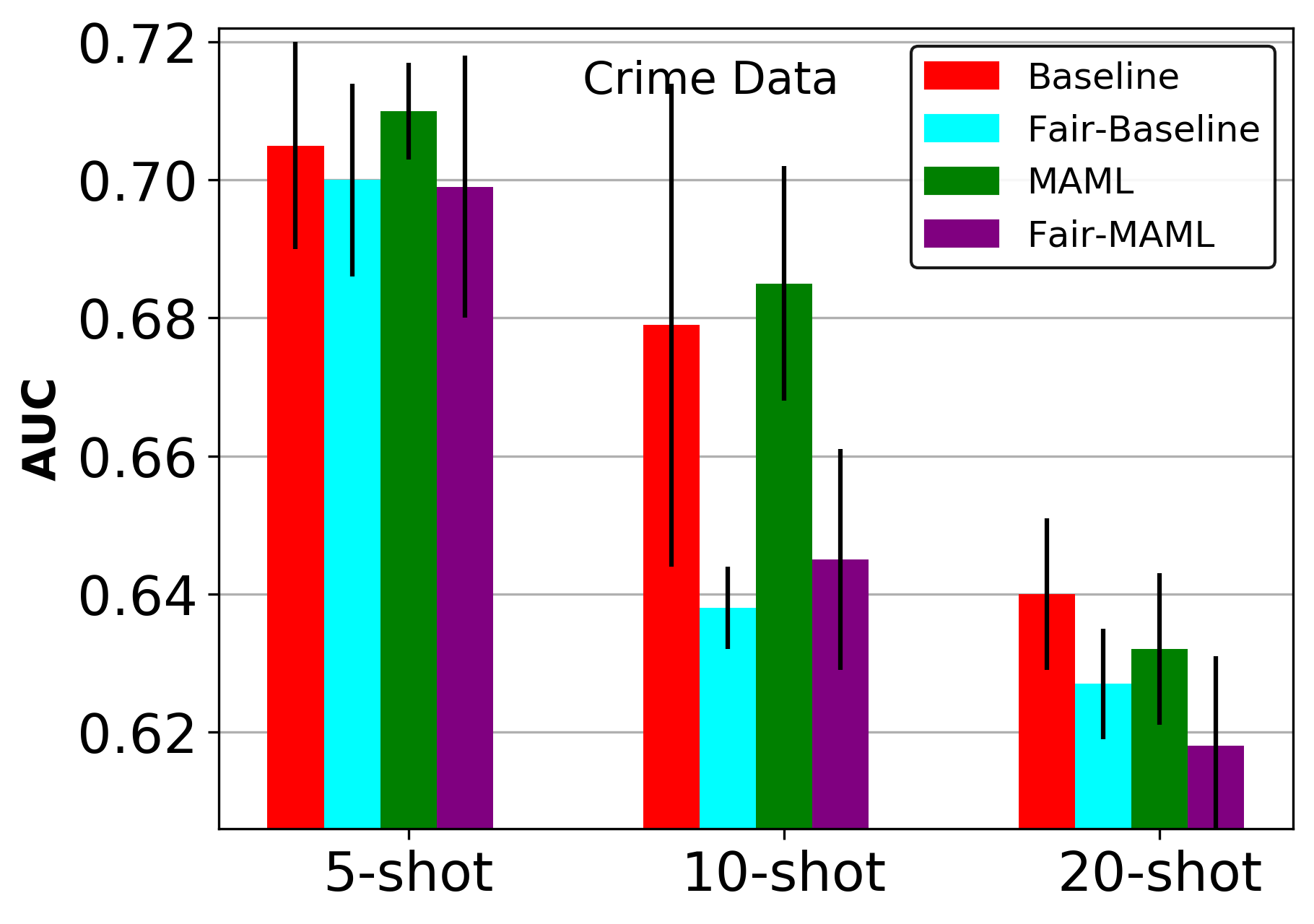}
        \caption{}
    \end{subfigure}
    \begin{subfigure}[b]{0.245\textwidth}
        \includegraphics[width=\textwidth]{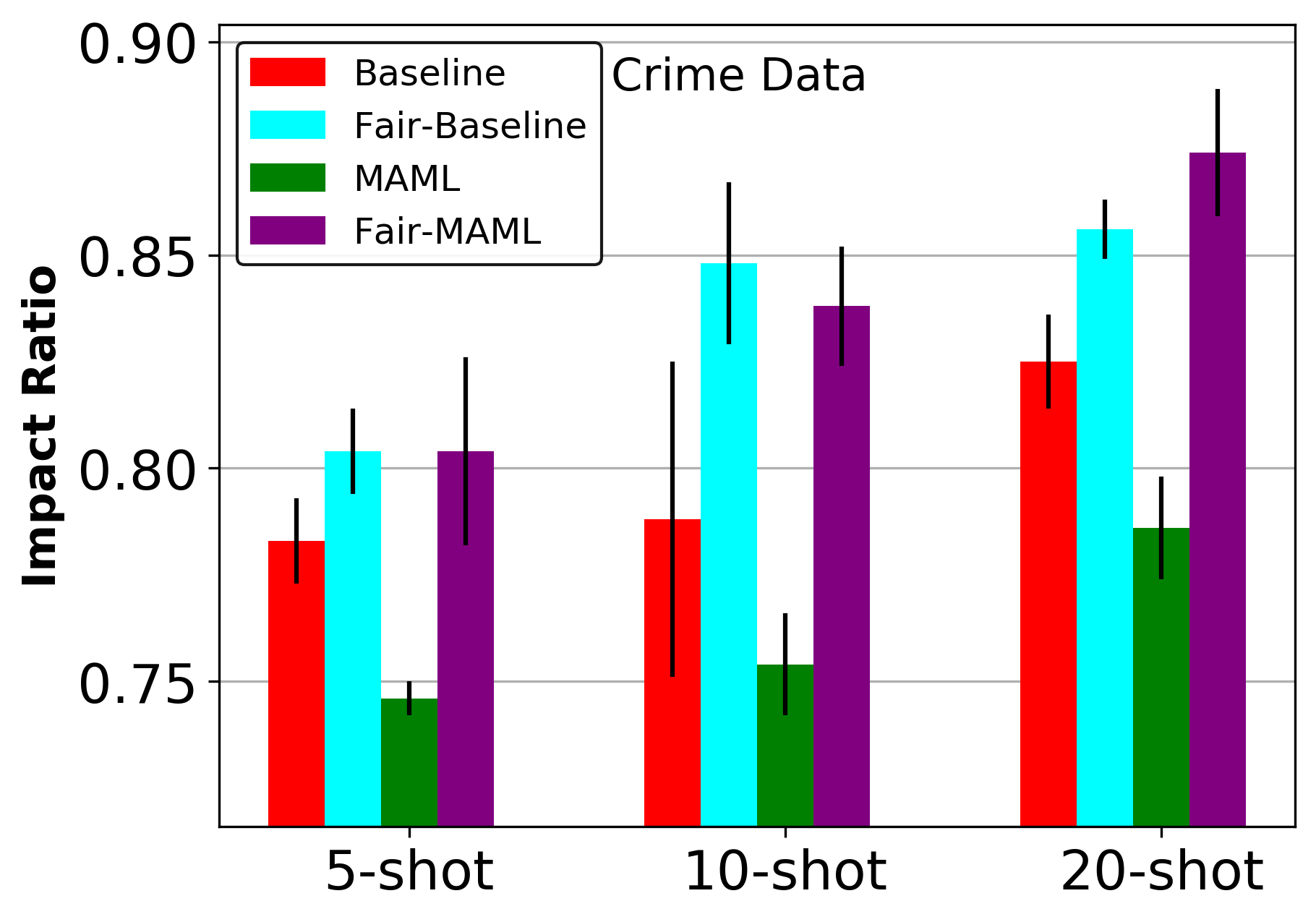}
        \caption{}
    \end{subfigure}
    
    
    \caption{Experiment results of controlling biases and fairness generalization to unseen data for meta-learning regression problem.}
    \label{fig:regression bar results}
\vspace{-5mm}
\end{figure*}

\textbf{Synthetic Data.} To evaluate, we start with a simple regression problem. We generated 12,000 synthetic data sets in total, 10,000 for training, 1000 for validation and 1000 for testing. Each data set can be considered as a single task. Specifically, for each data set, we generated 1000 data samples along with binary protected attributes uniformly. Each observation was uniformly assigned with a feature vector including seven explanatory attributes. Targets were generated from two Gaussian distributions with the same standard deviation of $\sigma=1$ but shifted means. To make each task unfair to some extent, targets from the unprotected group generated with arbitrary mean in $[0,10]$, but targets means from protected group increased by $[1,5]$ randomly.

The same \textbf{Chicago Communities and Crime (Crime)} data \cite{Zhao-ICDM-2019} that was applied in unfairness discovery experiments is continued to use for studying unfairness prevention in few-shot meta-learning model. Different from the previous setting, to simplify, we only keep one protected variable, \textit{``African American (\%)"}, and treat the rest (\textit{i.e. ``American Indian (\%)", ``Asian (\%)",} and \textit{``Other Ethnicity (\%)"}) as explanatory attributes. The Crime data set is divided into 801 sub-tasks according to different communities in the Chicago city. All tasks were further split into 501 for training, 100 for validation and 200 for testing. Each task contains 52 crime records. Since the feature in the original data that described the percentage of African American population is numeric, in this experiment, we convert it into binary values based on the majority ($>70\%$) population of Black and non-Black. Thus, each record represents a weekly information including 13 numeric explanatory variables and one binary protected variable.


All the attributes were standardized to zero mean and unit variance for all data sets and prepared for experiments. 
Our neural network trained follows the same architecture used by \cite{Finn-ICML-2017-(MAML)}, which contains 2 hidden layers of size of 40 with ReLU activation functions. When training, we use only one step gradient update with $2K$ samples of query set with a fixed learning rate of 0.01 and use Adam as the meta-optimizer. Similarly, we set the learning rate of 0.001 used to update the meta-loss in the outer loop. Hyperparameters are selected by held-out validation data. All experiments are repeated 10 times with the same settings. Results shown with these two methods in this paper are mean of experimental outputs followed by the standard deviation.

To evaluate performance, we fine-tune a single meta-learned model on varying numbers of $K\in\{5,10,20\}$ examples, and compare performance to two baselines: (a) the original MAML model \cite{Finn-ICML-2017-(MAML)}, and (b) the baseline method which pre-trains and entails training the network to regress to random functions and fine-tuning at the meta-testing stage using an automatically tuned step size. Both MAML and the baseline (pre-trained) models share the same neural network architecture and parameter settings. In order to distinguish our approach from the two baselines, we add a prefix ``Fair-" in Figure \ref{fig:syn loss/md iterations} and \ref{fig:regression bar results}.

In addition, we introduce two off-the-shelf \textbf{evaluation metrics} to measure biases. These measurements come into play of quantifying the extent of bias and are designed for indicating indirect discrimination. 

(1) \textbf{The area under the ROC curve (AUC)} \cite{Calders-ICDM-2013}. 
\begin{align}
    AUC = \frac{\sum_{(s_i,y_i)\in\mathcal{D}_+}\sum_{(s_j,y_j)\in\mathcal{D}_-}I(y_i>y_j)}{|\mathcal{D}_+|\times|\mathcal{D}_-|}
\end{align}
where $I(\cdot)$ is an indicator function which returns 1 if its argument is true, 0 otherwise. $AUC=0.5$ represents random predictability, thus $S$ is independent on $Y$. 

(2) \textbf{Impact Ratio (IR)} \cite{Pedreschi-SDM-2009}. 
\begin{align}
    IR = \frac{\sum_{y_i\in\mathcal{D}_+} y_i}{|\mathcal{D}_+|}\Big/\frac{\sum_{y_j\in\mathcal{D}_-} y_j}{|\mathcal{D}_-|}
\end{align}
It is defined as the ratio of mean over the protected and unprotected group in data $\mathcal{D}$. The decisions are deemed to be discriminatory if the ratio of positive outcomes for the protected attribute is below $80\%$ \cite{Biddle-Gower-2005}. $IR=1$ indicates that there is no bias of data $\mathcal{D}$.

\textbf{Experimental Results.} We evaluate the performance of our approach followed by \cite{Finn-ICML-2017-(MAML)} by fine-tuning the (Fair-) Baseline and models learned by (Fair-) MAML on $K=\{5,10,20\}$ data points. Results of validation loss and mean difference of each iteration are plotted in Figure \ref{fig:syn loss/md iterations}. In terms of losses (see Fig.\ref{fig:syn loss/md iterations} (a), (c) and (e)), MAML is outperformed than baseline methods and the gap between all methods is narrowing as the number of training data increases. Although our proposed approach (\textit{i.e.} Fair-Baseline and Fair-MAML) returns a bit bigger validation loss, this is due to the trade-off between fairness and accuracy. In Figure \ref{fig:syn loss/md iterations} (b), (d) and (f), our approach demonstrates success in controlling bias and decreasing validation losses, even trained with few-shot samples.

More experimental results with synthetic (see Figure\ref{fig:regression bar results}(a)-(d)) and real-world data set ((e)-(h)), as well as those are examined with two fairness evaluation metrics (AUC and IR) are shown in Figure \ref{fig:regression bar results}. Results through the Crime data demonstrate that our approach of controlling disparate treatment significantly decreases AUC and MD, and hence increases IR above the boundary of bias level of 80\% rules \cite{Biddle-Gower-2005} in contrast to methods without adding ``Fair-" constraints. Besides, the larger $K$ value (\textit{i.e.} more samples are considered in the support set), the better generalization capability of loss and fairness based on a few novel instances performs. This demonstrates the working efficiency of the proposed model.

MAML became a famous meta-learning algorithm because of its fast adaptation and good generalization performance on losses. However, our results showed it fails to control biases nor performs success in fairness generalization in a few-shot meta-learning. Our approach nevertheless makes up for this deficiency. 


\section{Conclusion and Future Work}
In this paper, for the first time we discover unfairness based on causal Bayesian network which reveals causal effect between all variables. In addition, we develop a novel algorithm based on risk difference in order to quantify the discriminatory influence for each protected variable in the graph. Besides, to prevent prediction from intervention of the protected variable, a fast-adapted bias-control approach by adding statistical parity constraints is proposed, which significantly mitigates dependence of prediction on the protected variable in each task and generalize both accuracy and fairness to unseen tasks. Due to the nature of MAML, which finds a task-specific model parameter for each task, one of the goal of future researches in few-shot learning is to design a fairness regulatory mechanism such that it automatically designs a task-specific fairness constraint through hyperparameter optimization techniques.

\section*{Acknowledgement}
This work is supported by the National Science Foundation (NSF) under Grant No \#1815696 and \#1750911.

\bibliographystyle{IEEEtran}
\bibliography{main}

\end{document}